\documentclass[letterpaper, 10 pt, journal, twoside]{IEEEtran}
\IEEEoverridecommandlockouts
\usepackage{cite}
\usepackage{amsmath,amssymb,amsfonts, dsfont}
\usepackage{algorithm, listings}
\usepackage{algpseudocode}	
\usepackage{graphicx}
\usepackage{textcomp}
\usepackage{xcolor}
\usepackage{mathtools}
\usepackage{blindtext, graphicx}
\usepackage{subcaption}
\usepackage{multirow}

\renewcommand{\vec}[1]{\boldsymbol{\mathbf{#1}}}
\newcommand{\mat}[1]{\boldsymbol{\mathbf{#1}}}
\def\argmin{\mathop{\arg\min}\limits}	%

\def\BibTeX{{\rm B\kern-.05em{\sc i\kern-.025em b}\kern-.08em
    T\kern-.1667em\lower.7ex\hbox{E}\kern-.125emX}}
    
\begin{document}
\markboth{IEEE Robotics and Automation Letters. Preprint Version. Accepted December, 2019} {Chen \MakeLowercase{\textit{et al.}}: SLOAM}

\title{SLOAM: Semantic Lidar Odometry and Mapping for Forest Inventory
\thanks{Manuscript received: September, 11, 2019; Revised November, 30, 2019; Accepted December, 23, 2019.}
\thanks{
This paper was recommended for publication by Editor Allison Okamura upon evaluation of the Associate Editor and Reviewers' comments. This work is supported by Trefo LLC under the NSF SBIR grant \#193856, by ARL grant ARL DCIST CRA 2911NF-17-2-0181, ONR grant N00014-07-1-0829, ARO grant W911NF-13-1-0350, INCT-INSac grants CNPq 465755/2014-3, FAPESP 2014/50851-0 and 2018/24526-5. This work is supported in part by the Semiconductor Research Corporation (SRC) and DARPA. We additionally thank NVIDIA for generously providing support through the NVAIL program.}
\thanks{
\IEEEauthorrefmark{1}Equal contribution.
$^{1}$Authors with GRASP Laboratory, University of Pennsylvania, United States. \{\texttt{\small chenste,elslee,quchao,
liuxu,kumar\}@seas.upenn.edu}. \newline
$^{2}$Authors with the Robot learning laboratory, ICMC - University of Sao Paulo, Brazil. \{\texttt{\small guinardari,rafrance\}@usp.br}}
\thanks{Digital Object Identifier (DOI): see top of this page.}
}

\author{
    \IEEEauthorblockN{Steven W. Chen$^{1}$\IEEEauthorrefmark{1}, Guilherme V. Nardari$^{2}$\IEEEauthorrefmark{1}, Elijah S. Lee$^{1}$, Chao Qu$^{1}$, Xu Liu$^{1}$, Roseli A. F. Romero$^{2}$, and Vijay Kumar$^{1}$}
}
\maketitle

\begin{abstract}
This paper describes an end-to-end pipeline for tree diameter estimation based on semantic segmentation and lidar odometry and mapping. Accurate mapping of this type of environment is challenging since the ground and the trees are surrounded by leaves, thorns and vines, and the sensor typically experiences extreme motion. We propose a semantic feature based pose optimization that simultaneously refines the tree models while estimating the robot pose. The pipeline utilizes a custom virtual reality tool for labeling 3D scans that is used to train a semantic segmentation network. The masked point cloud is used to compute a trellis graph that identifies individual instances and extracts relevant features that are used by the SLAM module. We show that traditional lidar and image based methods fail in the forest environment on both Unmanned Aerial Vehicle (UAV) and hand-carry systems, while our method is more robust, scalable, and automatically generates tree diameter estimations.   
\end{abstract}



\section{Introduction}
    \label{sec:intro}
\begin{figure}[t!]
\includegraphics[width=\columnwidth,trim={2cm 2cm 0cm 0cm},clip]{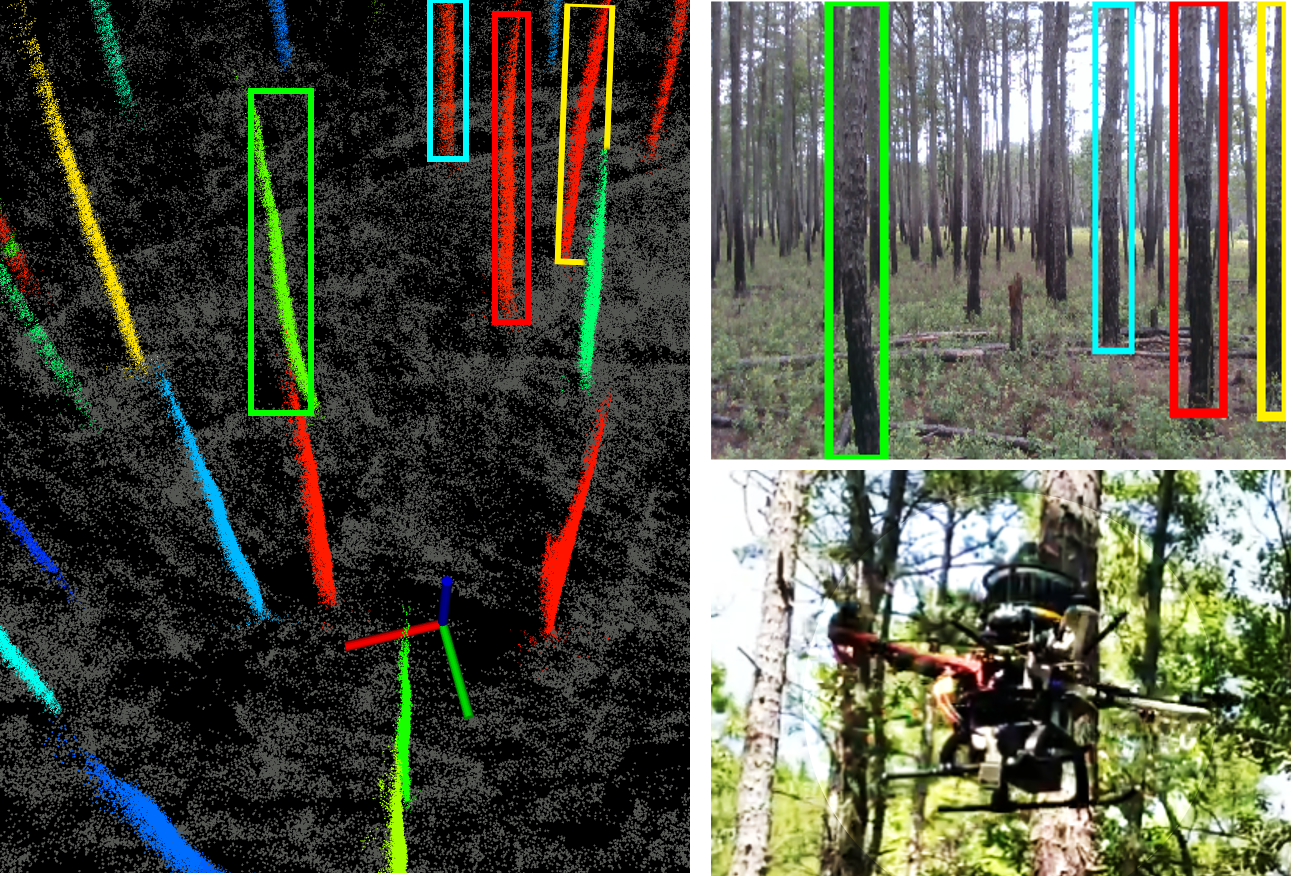}
\caption{Our UAV performing a timber inventory acquisition by simultaneously estimating its state and the tree diameters. \textcolor{black}{Bounding boxes are manually added to illustrate the same landmarks in the 3D point cloud and the 2D image.}}
\label{fig:robot_platform}
\end{figure}

\IEEEPARstart{O}{btaining} accurate timber inventory is crucial for forest managers. Foresters currently use Terrestrial Laser Scanning (TLS) sensors~\cite{DeConto2017, liang2016terrestrial} to extract tree metrics with high accuracy and reduced manpower~\cite{DeConto2017}. However, these sensors need to be moved around to measure the timber inventory at multiple locations for full coverage of the forest, requiring additional time and manpower. We propose a robotic timber cruise involving a robot navigating a stand of timber, measuring samples, and estimating the total forest volume. Fig.~\ref{fig:robot_platform} shows our Unmanned Aerial Vehicle (UAV) used in this work, and the resulting estimated timber map. This approach provides increased data granularity allowing foresters to make optimized management and harvesting decisions. 


Deploying UAV systems is appealing for a variety of applications including fruit counting~\cite{Chen2017, Liu2019}, disaster management~\cite{Lee2017}, and penstock inspection~\cite{8764006}. Previous works~\cite{perez2018architecture, Cui2016, maciel2018extending, Gao2019} focus on autonomous navigation in cluttered~\cite{Gao2019} or GPS-denied environments~\cite{perez2018architecture}. Many works employ UAVs in forests~\cite{Cui2016, maciel2018extending, Gao2019, Giusti2016, MacielPearson2019, Torresan2017, Puliti2015}. Most enable autonomous navigation by detecting and following forest trails using learning-based approaches~\cite{maciel2018extending, MacielPearson2019, Giusti2016}. Cui et al.~\cite{Cui2016} uses a 2D laser range finder to navigate autonomously, but makes the assumption that the UAV height does not drastically change. More recently,~\cite{Gao2019} develops a UAV platform equipped with a 3D lidar and IMU for navigating in timber environments and building point cloud maps, and~\cite{Torresan2017, Puliti2015} review the UAV applications in forestry. None of these prior works focus on the problem of estimating timber volume. 

Reche et al.~\cite{Reche2011} addresses the timber inventory problem by modeling trees from photographs, and Fritz et al.~\cite{Fritz2013} uses photogrammetric point clouds collected over canopy to map tree stems. Other works estimate tree attributes using TLS~\cite{DeConto2017, liang2016terrestrial}. 
Relying on photography or TLS can be time consuming, and capturing data over dense forest canopy may not provide accurate tree diameter at breast height (DBH).

Our work leverages UAVs and lidars for measuring timber inventory, particularly tree DBH estimation, by focusing on solving the simultaneous localization and mapping (SLAM) problem. Previous methods, such as lidar odometry and mapping (LOAM)~\cite{zhang2014loam}, rely on texture-based features that are brittle in forest environments. Our goal is to develop a Semantic LOAM (SLOAM) algorithm that extends ideas from semantic SLAM~\cite{bowman2017probabilistic} for forestry applications. Our key idea is utilizing a parameterized landmark shape representation to obtain more robust solutions to the SLAM and DBH problems.

We present an end-to-end pipeline for solving the DBH estimation problem. While our approach is specific to our primary forestry problem, \textcolor{black}{the framework and tools we develop can be generalized to other environments}. The pipeline consists of: (1) 3D point cloud labeling with virtual reality (VR); (2) range image segmentation with a fully convolutional neural network (FCN); (3) landmark instance detection with trellis graphs; (4) semantic lidar odometry; and (5) semantic lidar mapping. We first present the semantic lidar odometry and mapping modules, as they form the core ideas of the work. We then present the first 3 modules which relate to accurate detection of landmarks from individual lidar sensor readings. Finally, we present experimental results of our system in the field. The primary contributions of this paper are:
\begin{itemize}
\item A complete end-to-end pipeline for solving the forest DBH estimation problem using UAV systems; and 
\item SLOAM, a semantic framework that is capable of handling challenging environments with aggressive motions while simultaneously estimating landmark parameters.
\end{itemize}



\section{Problem Formulation}
    \label{sec:problem}
    \textbf{Primary Problem} (Forest DBH Estimation). Given a forest consisting of trees $\mathcal{L} \triangleq \{\mathcal{L}^{i}\}_{i=1}^{N}$, estimate the number of trees $N$, and the position and diameter at breast height (DBH) of each tree $\mathcal{L}^{i}$.

\textcolor{black}{The use of a mobile lidar sensor to solve the DBH estimation problem requires registering a sequence of lidar sweeps into a common reference frame.} This registration typically involves estimating the trajectory of the lidar sensor over time, which leads to the following problem:

\textbf{General Problem} (Semantic LOAM). Given a sequence of lidar sweeps $\mathcal{P} \triangleq \{\mathcal{P}_{k}\}_{k=1}^{K}$ collected in an environment consisting of objects $\mathcal{L} \triangleq \{\mathcal{L}^{i}\}_{i=1}^{N}$, estimate the sensor state trajectory $\mathcal{X}$, the number of objects $N$, and the model parameters and classes of each object $\mathcal{L}^{i}$.

Under specific assumptions on the model parameterization and class of objects $\mathcal{L}^{i}$, a solution to the general Semantic LOAM problem will yield a solution to the primary DBH Estimation problem. 
    
    
\section{Semantic Lidar Odometry and Mapping}
	\label{sec:odometry_and_mapping}

The purpose of the lidar odometry algorithm is to estimate the rigid 6-DOF motion of the lidar within a $360^\circ$ lidar sweep. The purpose of the lidar mapping algorithm is to estimate the 6-DOF pose of the lidar in the world, and register the point cloud in the world frame. We follow the framework introduced in the original LOAM paper~\cite{zhang2014loam}. SLOAM incorporates semantic landmark features in an end-to-end system to increase robustness, scalability, and performance in challenging environments where LOAM fails, such as forests. 

SLOAM relies on landmark models and features to estimate motion and register the lidar point clouds. The choice of which landmark features to use, and how to model the landmarks, depends on the environment. In forests, good landmark features are trees and ground modeled as cylinders and planes. We will assume that these features have been segmented and clustered into individual tree instances by an external segmentation and instance detection module described in Sec.~\ref{sec:robust_tree_and_Ground}.

For the previous lidar sweep  $\mathcal{P}_{k}$ that started with time stamp $t_{k}$, let $\mathcal{L}_{k} \triangleq \{\mathcal{L}_{k}^{i}\}_{i=1}^{N_{k}}$ be the set of tree landmark models of size $N_{k}$. For a given $\mathcal{L}_{k}^{i}$, let $\mathcal{T}_{k}^{i} \triangleq \{\mathbf{p}_{j}\}_{j=1}^{\delta_{i,k}}$ be the set of tree feature points of size $\delta_{i,k}$ associated with the landmark, and let $\mathcal{T}_{k} \triangleq  \{\mathcal{T}_{k}^{i}\}_{i=1}^{N_{k}}$ be the aggregation of all tree features. Let $\mathcal{G}_{k} \triangleq \{\mathbf{p}_{l}\}_{l=1}^{\gamma}$ be the set of ground feature points of size $\gamma$. Denote $\overline{\mathcal{L}}_{k}, \overline{\mathcal{T}}_{k}, \overline{\mathcal{G}}_{k}, \overline{\mathcal{P}}_{k}$ as the projections of $\mathcal{L}_{k}$, $\mathcal{T}_{k}$, $\mathcal{G}_{k},\mathcal{P}_{k}$ to the lidar frame at time stamp $t_{k+1}$. Let $t$ be the current time stamp, and $\mathbf{T}^{L}_{k+1} = [t_{x}, t_{y}, t_{z}, \theta_{x}, \theta_{y}, \theta_{z}]$ be the lidar pose transform between $[t_{k+1}, t]$, where $t_{x}, t_{y}, t_{z}$ are the translations along the $x$, $y$, and $z$ axes in the lidar frame $\{L\}$, and $\theta_{x}$, $\theta_{y}$, and $\theta_{z}$ are the rotation angles following the right-hand rule. Let $\mathbf{T}^{W}_{k+1}$ be similarly defined in the world frame $\{W\}$, which defines the sensor state trajectory $\mathcal{X} \triangleq (\mathbf{T}^{W}_{k})_{k=1}^{K}$. Both the semantic odometry and mapping algorithms estimate the pose transforms by performing a data association between tree and ground features in the current sweep with tree and ground models in a previous sweep (odometry) or map (mapping).


\subsection{Model Parameterization and Distance Functions}

We model the ground locally as a plane parameterized by $\vec{\pi} = (\vec{\omega}, \beta)$, where $\vec{\omega}$ is the normal of the plane, and $\beta$ is the offset such that the plane is defined by $\{\vec{x} | \langle\vec{x}, \vec{\omega}\rangle + \beta = 0\}$. Given a point $\vec{p}$ and plane $\vec{\pi}$, let $\vec{x}_{0}$ be a point on the plane. We can then define a point to plane distance:

\begin{equation}
    \label{eq:point_to_plane_distance}
    d_{\pi}(\vec{\pi}, \vec{p}) = \frac{\langle -(\vec{p} - \vec{x}_{0}), \vec{\omega} \rangle}{||\vec{\omega}||}.
\end{equation}

Since the ground is relatively flat and lidar upright, ground features provide constraints in the $z$ axis, but not the $x$ and $y$ axes, and we need to also detect tree models and features.  

We parameterize the tree cylinder models following the method in~\cite{lukacs1997geometric}. Let $s = (\rho, \phi, \nu, \alpha, \kappa)$ be the parameters of a cylinder model. \textcolor{black}{To understand these parameters, let $\vec{n}$ be the cylinder normal at the point on the cylinder closest to the origin, and $\vec{a}$ be the axis of the cylinder}. The first parameter $\rho$ is the distance from the origin to the closest point on the cylinder, $\phi$ is the angle between the projection of $\vec{n}$ onto the $xy$ plane with the $x$ axis, $\nu$ is the angle between $\vec{n}$ and the $z$ axis, $\alpha$ is the angle between $\vec{a}$ and the partial derivative of $\vec{n}$ with respect to $\nu$, and $\frac{1}{\kappa}$ is the radius of the cylinder. \textcolor{black}{For an illustration of the cylinder model parameters and angles, please refer to~\cite[Fig.~3]{lukacs1997geometric}}. More specifically, given $(\rho, \phi, \nu, \alpha, \kappa)$, we can compute the normal $\vec{n}$ and axis $\vec{a}$ as:

\begin{equation}
\label{eq:polar_parameterization}
\begin{aligned}
\mathbf{n} &= (\cos\phi \sin\nu, \sin\phi \sin\nu, \cos\nu) \\
\mathbf{a} &= \mathbf{n}^{\nu} \cos \alpha + \overline{\mathbf{n}^{\phi}}\sin \alpha \\
\end{aligned}
\end{equation}
where $\mathbf{n}^{\nu} = (\cos \phi \cos \nu, \sin \phi \cos \nu, -\sin \nu)$ is the partial derivative of $\mathbf{n}$ w.r.t $\nu$, and $\overline{\mathbf{n}^{\phi}} = \frac{\mathbf{n}^{\phi}}{\sin \nu} = (-\sin \phi, \cos \phi, 0)$ is derived from the partial derivative of $\mathbf{n}$ w.r.t $\phi$. 

The point to cylinder distance between a point $\vec{p}$ and cylinder $\vec{s}$ is:

\begin{equation}
\label{eq:true_point_to_cylinder_distance}
d_{s}(\vec{s}, \vec{p}) = \Bigl|(\vec{p} - (\rho + \frac{1}{\kappa})\vec{n}) \times \vec{a}\Bigr| - \frac{1}{\kappa}.
\end{equation}

\subsection{Estimating Tree and Ground Models}

Given the set of feature points $\mathcal{T}_{k+1}^{i}$ of size $\delta_{i, k+1}$ corresponding to landmark $\mathcal{L}_{k+1}^{i}$, the geometric least squares approach to estimating the parameters of a cylinder $s$ is to stack the point to cylinder distances for each feature point and optimize over the parameters of $s$. However, the point to cylinder distance can encounter singularities when the curvature $\kappa$ decreases, which introduces difficulties in the numerical optimization procedure~\cite{lukacs1997geometric}. As a result, for a feature $\mathbf{p}$ and cylinder $\vec{s}$, we approximate the point to cylinder distance with the following distance which has the same zero set and derivatives at the zero set as the true distance function, but additionally behaves well when the curvature $\kappa$ decreases:

\begin{equation}
\label{eq:approximate_point_to_cylinder_distance}
\hat{d}_{s}(\mathbf{s}, \mathbf{p}) = \frac{\kappa}{2}(|\mathbf{p}|^{2} - 2\rho \langle \mathbf{p}, \mathbf{n}\rangle - \langle \mathbf{p}, \mathbf{a}\rangle^{2} + \rho^{2}) + \rho - \langle \mathbf{p}, \mathbf{n}\rangle).
\end{equation}

The geometric least squares optimization problem is then solved as follows:

\begin{equation}
\label{eq:geometric_least_squares}
\begin{aligned}
& \argmin_{\rho, \phi, \nu, \alpha, \kappa}
& & \sum_{j=0}^{\delta_{i,k+1}}\hat{d}_{s}(\mathbf{s}, \mathbf{p}_{j})\\
\end{aligned}
\end{equation}
It is easy to ensure that $\kappa, \rho \geq 0$, by optimizing over $\sqrt{\kappa}, \sqrt{\rho}$. Thus given a set of tree features $\mathcal{T}_{k+1}^{i}$, solving Prob.~\eqref{eq:geometric_least_squares} will yield a cylinder landmark model $s$.

Our approach to estimate the local ground plane models is similar to previous works~\cite{shan2018lego, zhang2014loam}. However, the challenge with forest environments is that the ground is frequently covered by brush and small shrubs that are irregularly shaped. While we describe a ground segmentation method that tries to filter out this noise in Sec.~\ref{sec:robust_tree_and_Ground}, it is extremely challenging to completely filter out all of the underbrush in natural forest environments. This noise can be problematic for standard ground plane estimation methods used in methods such as \cite{shan2018lego} which only utilize $3$ points to estimate the plane. 

Instead, we use the following robust ground estimation method. Given a set of $m$ ground feature points denoted as the $3 \times m$ matrix $\mat{G}$, we use Singular Value Decomposition (SVD) to estimate a model of the ground plane. Let $\overline{\mat{G}}$ be the centroid, and $\underline{\mat{G}} = \mat{G} - \overline{\mat{G}}$ be the mean shifted ground feature points. We can then compute the SVD of $\underline{\mat{G}} = U\Sigma V^{T}$ where $U \in \mathbb{R}^{3\times3}$, $\Sigma \in \mathbb{R}^{3 \times m}$ and $V \in \mathbb{R}^{m \times m}$. The estimate of the normal $\vec{\omega}$ will be the left singular vector of $U$ corresponding to the least singular value. Given $\vec{\omega}$ and the centroid $\overline{\mat{G}}$, we can compute $\beta$ by using the equation of the plane.

\subsection{Motion Estimation and Data Association}

Given a set of tree features $\mathcal{T}_{k+1}$ and ground features $\mathcal{G}_{k+1}$ in the current sweep $\mathcal{P}_{k+1}$, as well as tree features $\overline{\mathcal{T}}_{k}$, ground features $\overline{\mathcal{G}_{k}}$, and tree landmark models $\overline{\mathcal{L}_{k}}$ from the previous sweep $\mathcal{P}_{k}$ projected to $t_{k+1}$, we can estimate the pose transform $\mathbf{T}_{k+1}^{L}$ by optimizing the following non-linear least squares problem:
\begin{equation}
    \label{eq:pose_optimization}
    \begin{aligned}
& \argmin_{\mathbf{T}_{k+1}^{L}}
& & \lambda_{t}\sum_{i=1}^{N_{k+1}}\sum_{j=0}^{\delta_{i,k+1}}d_{s}(\mathbf{s}_{j}, \mathbf{p}_{j}) + \lambda_{g}\textcolor{black}{\sum_{l=0}^{\gamma}d_{\pi}(\mat{\pi}_{l}, \mat{p}_{l})},\\
\end{aligned}
\end{equation}
\textcolor{black}{where $\lambda_{t} = \frac{\gamma}{\sum_{i=1}^{N_{k+1}}\delta_{i,k+1}}$ and $\lambda_{g} = \frac{1}{\lambda_{t}}$ balance the frequency between tree and ground features}. In this optimization problem, we use the true point to cylinder distance defined in Eqn.~\eqref{eq:true_point_to_cylinder_distance}, as the singularity from $\kappa$ is eliminated since $\kappa$ is now a constant. Notice that each of the distance function $d_{s}$ ($d_{\pi}$) depends on associating each feature point $\vec{p}_{j}$ ($\vec{p}_{l}$) with an object model $\vec{s}_{j}$ ($\vec{\pi}_{l}$), which is called data association. This data association process is different for the tree features and ground features.

For a tree feature $\vec{p}_{j} \in \mathcal{T}_{k+1}^{i}$, we need to find its landmark correspondence in $\overline{\mathcal{L}}_{k}$. We first project the feature $\mathbf{p}_{j}$ taken at time $t$ to $\mathbf{p}_{j}'$ in the frame at time $t_{k+1}$ using an initial guess of $\vec{T}_{k+1}^{L}$ at time $t$. There are two methods to perform the data association for tree features. In the first method, we match each feature $\vec{p}_{j}$ to the model in $\overline{\mathcal{L}}_{k}$ with the smallest orthogonal distance to $\mathbf{p}_{j}'$. In the second method, we find the nearest neighbor of $\mathbf{p}_{j}'$ in $\overline{\mathcal{T}}_{k}$. Since each feature point in $\overline{\mathcal{T}}_{k}$ is associated with a cylinder model in $\overline{\mathcal{L}}_{k}$, we then perform data association by matching $\mathbf{p}_{j}$ to the cylinder associated with the nearest neighbor. 

The first data association method is desirable, especially during the lidar mapping algorithm, because it only needs to maintain $\overline{\mathcal{L}}_{k}$, whereas the second method needs to maintain both $\overline{\mathcal{L}}_{k}$ and $\overline{\mathcal{T}}_{k}$. In a global map, the size of $\overline{\mathcal{L}}_{k}$ is independent of the number of sweeps seen, whereas the size of $\overline{\mathcal{T}}_{k}$ will grow over time. However, the second method can be more accurate since it preserves more detailed information through the features at the expense of having a map representation that increases with the number of sweeps. In this work, our main concern is to first obtain accuracy and robustness in forest environments, so we decide to employ the second method for data association. However, it is easy to switch to the first method when scaling to large environments. 

\textcolor{black}{Both data association methods are general to other environments beyond forests. The first method only requires a convex representation of the landmark models in order to efficiently compute a distance function, while the second method can be applied more generally as it directly compares features to features. As a result, our proposed data association methods extend beyond our primary problem of Forest DBH Estimation to the more general Semantic LOAM problem.}

SLOAM handles the ground feature points differently than the tree feature points. For a given ground feature point $\vec{p}_{l} \in \mathcal{G}_{k+1}$ at time $t$, we project it back to $\vec{p}_{l}'$ at time stamp $t_{k+1}$. We then find the set of nearest neighbors to $\vec{p}_{l}'$ in $\overline{\mathcal{G}}_{k}$, and compute the ground plane corresponding to this set using the SVD. We thus compute a ground plane for each feature $\vec{p}_{l}$, rather than explicitly maintaining and updating the plane model parameters as we do for the  tree cylinders. Our treatment of the ground plane is thus similar to~\cite{shan2018lego}. 


\subsection{Semantic Lidar Odometry and Mapping}
\begin{algorithm}[t!]
	\caption{Semantic Lidar Odometry}
	\begin{algorithmic}[1]
        \State \textbf{input} : $\overline{\mathcal{L}}_{k}, \overline{\mathcal{T}}_{k}, \overline{\mathcal{G}}_{k}$ from $\overline{\mathcal{P}}_{k}$
        \State \quad \quad \quad $\mathcal{T}_{k+1}, \mathcal{G}_{k+1}$ tree and ground features from $\mathcal{P}_{k+1}$
        \State \quad \quad \quad $\vec{T}_{k+1}^{L}$ initial pose transform from last recursion
        \State \textbf{output} : $\overline{\mathcal{L}}_{k+1}, \overline{\mathcal{T}}_{k+1}, \overline{\mathcal{G}}_{k+1}$, newly computed $\vec{T}_{k+1}^{L}$
        \If{at the beginning of a sweep $\mathcal{P}_{k+1}$} 
            \State $\vec{T}_{k+1}^{L} \gets \vec{0}$
            \State $\mathcal{L}_{k+1} = \emptyset$ 
        \EndIf
                \For{each tree instance $\mathcal{T}_{k+1}^{i}$} 
                \State Update tree models $\mathcal{L}_{k+1}$ by \label{alg:tree_update} solving~\eqref{eq:geometric_least_squares}
                \EndFor
                \For{each tree instance $\mathcal{T}_{k+1}^{i}$} \label{alg:common_steps_first}
                    \For{each tree point}
                        \State Find a tree model in $\overline{\mathcal{L}}_{k}$ as the 
                        \State \quad correspondence, then compute a point to 
                        \State \quad cylinder distance based on~\eqref{eq:approximate_point_to_cylinder_distance} and stack
                        \State \quad the equation to~\eqref{eq:pose_optimization}
                    \EndFor
                \EndFor
                \For{each ground point}
                    \State Find a ground plane model as the 
                    \State \quad correspondence, then compute a point to 
                    \State \quad plane distance based on~\eqref{eq:point_to_plane_distance} and stack
                    \State \quad the equation to~\eqref{eq:pose_optimization}
                \EndFor
                \State Update $\vec{T}_{k+1}^{L}$ by solving~\eqref{eq:pose_optimization} \label{alg:common_steps_last}
            \If{at the end of a sweep $\mathcal{P}_{k+1}$}
                \State Project $\mathcal{L}_{k+1}$ to $t_{k+2}$ to form $\overline{\mathcal{L}}_{k+1}$
                \State Project $\mathcal{T}_{k+1}, \mathcal{G}_{k+1}$ to $t_{k+2}$ to form $\overline{\mathcal{T}}_{k+1}, \overline{\mathcal{G}}_{k+1}$
                \State \Return $\overline{\mathcal{L}}_{k+1}, \overline{\mathcal{T}}_{k+1}, \overline{\mathcal{G}}_{k+1}, \vec{T}_{k+1}^{L}$
            \Else
                \State \Return $\vec{T}_{k+1}^{L}$
            \EndIf
	\end{algorithmic}
	 \label{alg:semantic_lidar_odometry}
\end{algorithm}

The semantic lidar odometry algorithm is presented in Alg.~\ref{alg:semantic_lidar_odometry}. \textcolor{black}{Depending on implementation, multiple recursions can be executed with growing $\mathcal{T}_{k+1}$ and $\mathcal{G}_{k+1}$ as the sweep $\mathcal{P}_{k+1}$ begins from $t_{k+1}$ and ends at $t_{k+2}$, or it can be executed just once per sweep}. It follows the overall framework of the original lidar odometry algorithm in~\cite{zhang2014loam}, except with the incorporation of semantic models and features. 

The semantic lidar mapping algorithm has analogous inputs and outputs to the lidar odometry algorithm. It takes as input $\overline{\mathcal{L}}_{k+1}, \overline{\mathcal{T}}_{k+1}, \overline{\mathcal{G}}_{k+1}, \vec{T}_{k+1}^{L}$, which is the output of the lidar odometry algorithm. It then follows the similar steps in Alg.~\ref{alg:semantic_lidar_odometry} lines~\ref{alg:common_steps_first}-~\ref{alg:common_steps_last} to estimate $\mat{T}_{k+1}^{W}$. Rather than comparing the new features to $\overline{\mathcal{L}}_{k}, \overline{\mathcal{T}}_{k}, \overline{\mathcal{G}}_{k}, \vec{T}_{k+1}^{L}$, the mapping algorithm compares them to its map parameters $\mathcal{L}_{k}^{W}, \mathcal{T}_{k}^{W}, \mathcal{G}_{k}^{W}, \vec{T}_{k+1}^{W}$.

The main difference is initializing and updating these map parameters. At time $t=0$, the set of map landmarks is empty $\mathcal{L}_{k}^{W} = \emptyset$. At each subsequent sweep $\mathcal{P}_{k+1}$, we combine the odometry output $\overline{\mathcal{L}}_{k+1}$ with $\mathcal{L}_{k}^{W}$. For each tree feature in the odometry output $\overline{\mathcal{T}}_{k+1}$, we assign it to a tree cylinder in $\mathcal{L}_{k}^{W}$ using the same data association process as in the lidar odometry algorithm. However, if a feature point is not close to any tree cylinders, we mark it as an unassigned point. After performing this process, we check the unassigned points and determine which belong to the same tree in the odometry output $\overline{\mathcal{L}}_{k+1}$. If a large enough number of points belong to the same tree in $\overline{\mathcal{L}}_{k+1}$, we add that tree to $\mathcal{L}_{k}^{W}$, taking care to project the tree cylinder model into the world frame. 

The process of aggregating the tree and ground features is the same as in the original LOAM algorithm~\cite{zhang2014loam}. We can thus combine the semantic lidar odometry and mapping algorithms to estimate the global pose transform $\mat{T}_{k+1}^{W}$ as well as accumulate both the feature points $\mathcal{T}_{k}^{W}, \mathcal{G}_{k}^{W}$, and the tree landmark models $\mathcal{L}_{k}^{W}$. The explicit estimation of $\mathcal{L}_{k}^{W}$ allows us to automatically retrieve the number of trees, along with an estimate of their radius, at the end of SLOAM. It thus demonstrates how a solution to SLOAM will yield a solution to our primary DBH Estimation problem.

\section{Robust Tree and Ground Detection}	
    \label{sec:robust_tree_and_Ground}

Sec.~\ref{sec:odometry_and_mapping} assumes access to the tree features $\mathcal{T}_{k+1}$ and ground features $\mathcal{G}_{k+1}$, and its performance is highly dependent on the quality and reliability of these features. However, obtaining these features is challenging in itself, and we present a 3 part module that reliably extracts tree features, as well a specialized segmentation method to extract ground features. These 3 modules for tree feature extraction can be generalized to other semantic features as well, and as a result can be combined with the SLOAM framework as a full end-to-end system for general environments. 


\subsection{Virtual Reality Point Cloud Labeling}

\textcolor{black}{Our data-driven deep learning segmentation method requires a large training data set in order to obtain good performance and generalization. However, there are few data sets for lidar point clouds in forests, and we need to collect and label our own data. Most current labeling tools use mouse and keyboard and are designed for 2D images, which are ill-suited for labeling 3D point clouds.}


\textcolor{black}{Virtual reality has frequently been used to visualize point clouds~\cite{bruder2014poster}, and recently to label them~\cite{stets2017visualization}. However, most of these applications and tools have been developed in the graphics community, and are either not suitable nor easy to adapt to a robotics application. As a result, we designed a custom VR labeling tool that interfaces with ROS in order to obtain and label point clouds from standard mobile lidar platforms. The VR labeling tool is programmed using the Unity video game engine with the Oculus Rift. }

\begin{figure}[t!]
\centering
\includegraphics[width=\columnwidth, trim={0cm 0cm 0cm 4cm},clip]{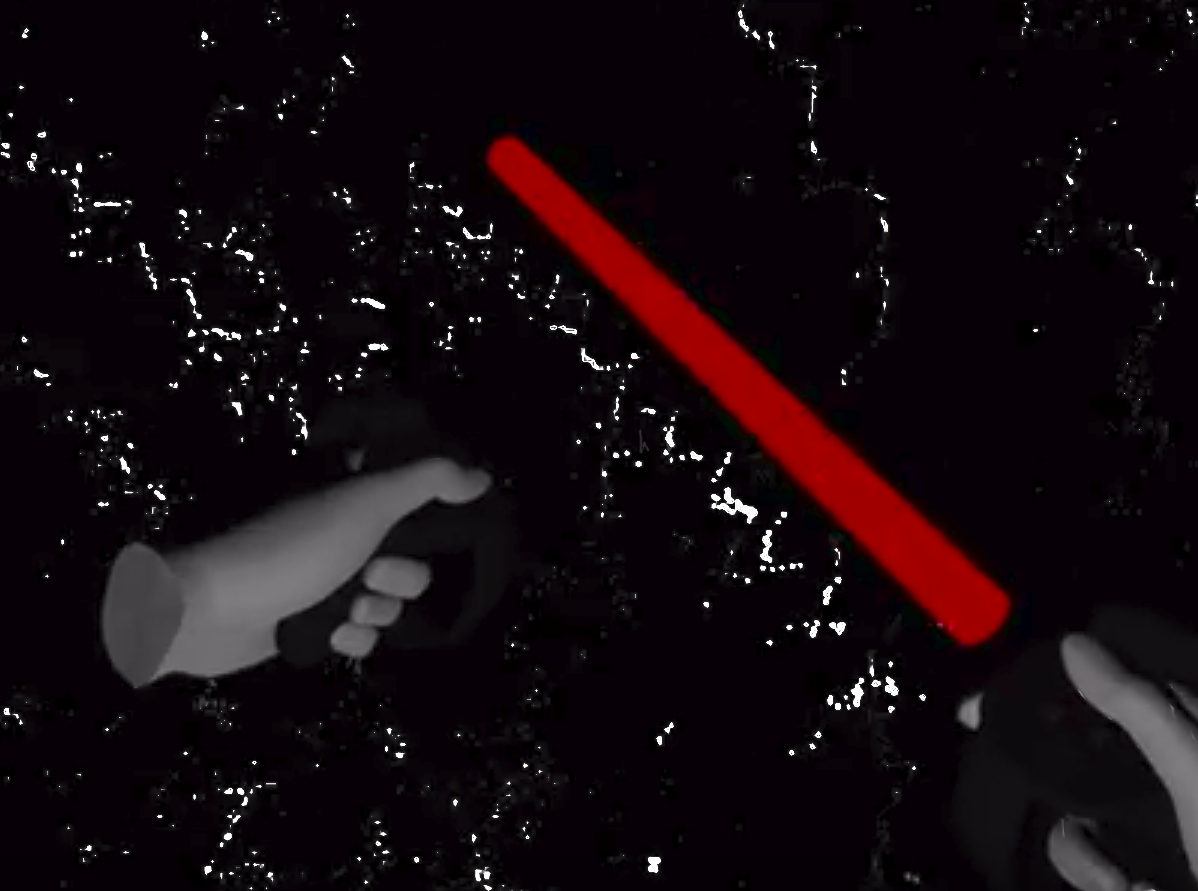}
\caption{Oculus Rift virtual reality point cloud labeling tool. The white points represent the point cloud, the red cylinder is the labeling primitive, and the gray hands represent the virtual hands of the labeling user.}
\label{fig:VR_tool}
\end{figure}

\textcolor{black}{In order to increase tree labeling efficiency, we design a labeling cylinder primitive, and other similar primitives can be used to label other objects. As shown in Fig.~\ref{fig:VR_tool}, the user labels points by placing the labeling primitive over the points. The user is able to easily rotate, translate and scale the point cloud using these handsets. After all tree points have been successfully labeled, the user can easily store the point cloud and cylinders into a database and proceed to the next point cloud without leaving the labeling environment, thus facilitating large acquisitions of labels.}

\subsection{Deep Learning Tree Segmentation}
Natural forests feature large levels of occlusion and noise, and it is challenging to reliably segment lidar tree points at large scale~\cite{tremblay2019automatic}. Most approaches use clustering-based methods~\cite{pierzchala2018mapping}, which are limited to relatively clean forests. As a result, our approach is to use a deep neural network for point cloud segmentation to extract the tree features $\mathcal{T}_{k+1}$. 


Rather than directly operating on the point cloud and requiring the use of slower network architectures such as PointNet++~\cite{qi2017pointnet++}, our segmentation network takes as input a 2D range image of size $h \times w$ where $h$ is the number of beams and $w$ is the number of azimuth readings of the sensor. Operating on the range image opens up the use faster FCN architectures. \textcolor{black}{Prioritizing speed, we use a simplified version of ERFNet~\cite{romera2017erfnet} with Non-bottleneck-1D structure, Downsampler and Deconvolution layers~\cite[Sec. IIIb]{romera2017erfnet}. }


The input is constructed by representing every 3D point as a pixel in the range image according to its position and the lidar sensor model. This representation captures spatial relationships through convolution operations and is more computationally efficient.
One drawback is that far apart points with similar altitude and azimuth angles, but different depth ranges, will be neighbors in the range image, and as a result can be challenging to differentiate. To minimize this effect, during training and inference we sample the point cloud according to a radius range relative to the sensor. The segmentation network will output a predicted segmentation for each sampled image, and the final network prediction is given by the sum of all the sampled predictions. This output is a mask assigning semantic labels for every pixel which are projected back to the corresponding 3D points in the lidar point cloud. 

\begin{figure}[t!]
\centering
\includegraphics[width=\columnwidth]{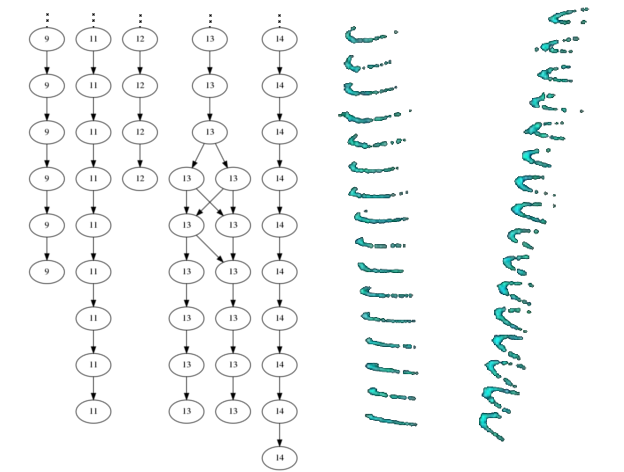}
\caption{\textbf{Left}: Trellis graph with 5 detected trees (lower beams at top, higher beams at bottom). Tree 13 exhibits a fork structure. \textbf{Right}: Detected tree points. Each of the 16 beams of our VLP-16 lidar passes over the trees, highlighting the lidar output structure and connection to trellis graphs.}
\label{fig:trellisExample}
\end{figure}



\subsection{Ground Segmentation}
We next perform ground segmentation to extract $\mathcal{G}_{k+1}$. Rather than a neural network, we use simple heuristics due to their effectiveness and simplicity. We make the basic assumption that the ground is locally planar, but not necessarily globally planar. In a given scan, we first remove points that the neural network determines to be part of a tree. As a result, we are left with points that belong either to the ground, shrub, or leaves. The core idea is that the ground should appear below all of these other points. However, we cannot simply extract the lowest points in $z$ from the lidar sweep, as this approach will fail if the terrain is sloped or features other complicated behavior. We instead divide the lidar sweep into a circular grid specified by the distance and angle of the space around the sensor. Within each grid cell, we retain the lowest points in the $z$ direction, where the number of points to retain is a hyperparameter. By retaining fewer points, it is more likely that the points in $\mathcal{G}_{k+1}$ are actually ground points, but the total number of ground features $\gamma$ will be smaller.

\subsection{Instance Detection}
\label{ssec:instance_detection}
Once we segment the tree points in the lidar sweep, we next group them into individual tree landmarks $\mathcal{T}_{k+1}^{i}$. The lidar output has a natural structure as exhibited in the range image, specifically that points can be sorted by beam and arranged by the rotation direction of the lidar. This output can be viewed as a trellis graph~\cite{forney1973viterbi}, where each slice of the trellis represents a single lidar beam. We exploit this trellis structure to detect tree instances. The key observation motivating our approach is that in the presence of gravity, a higher beam detecting an object indicates that a lower beam will likely also detect that same object. As a result, in the trellis graph, an object will appear as a path that starts from earlier slices (lower beams) and ends at later slices (higher beams). By properly constructing a weighted trellis graph, we can identify tree instances by efficiently solving for shortest routes using the dynamic programming Viterbi Algorithm~\cite{forney1973viterbi}, or even more simply a greedy algorithm. Since this approach only assumes gravity, it generalizes to other objects besides trees.

The vertices of the trellis graph are clusters of points belonging to the same object in a beam, and the edges are the connections between these vertices across beams. \textcolor{black}{We assume that points on the same object have similar depth readings, and construct vertices in the trellis graph by clustering contiguous points within a single beam according to a depth threshold. We create the edges by connecting the vertices between previous and succeeding slices. To reduce computation, we then remove edges where the distance between the centroid of the points in each vertex exceeds a distance threshold, as those vertices are unlikely to belong to the same tree. We specify the weights of each edge by constructing a score function, such as the distance between the vertex centroids.}

Once we have specified the trellis graph, we greedily find the tree instances by starting from vertices in the early slices and finding shortest routes through the trellis. We add the route as a tree instance if it exceeds a minimum path length threshold indicating that the tree was detected by enough lidar beams, and if the total path cost is less than a path weight threshold, indicating that the centroids of each vertex in the tree are close together. Although these shortest routes can be solved optimally with the Viterbi algorithm, we found that a greedy approach of following the minimum-weight outgoing edges at each vertex worked for tree instances.

There are $2$ auxiliary benefits of using this trellis approach. First, it immediately provides an initial guess for the cylinder parameters $(\rho, \phi, \nu, \alpha, \kappa)$. Since the geometric least squares Prob.~\eqref{eq:geometric_least_squares} is non-linear, it is extremely important to obtain a good intialization. We define the focus point of a vertex as the mean of the two points that are furthest from each other in the vertex. The focus point is an estimate of the center of the circle defining the shape of an individual vertex. The radius of this circle can be used to initialize the radius of the tree, and an Orthogonal Distance Regression (ODR) through the focus points of a tree can be used to estimate the axis $\vec{a}$ and normal $\vec{n}$ of the cylinder model of the tree. The second benefit of this trellis approach is that it can identify other properties of trees, for example forks in the tree where a single trunk branches into two parts. \textcolor{black}{Fig.~\ref{fig:trellisExample} displays two trees detected in an organized lidar scan. It also depicts a portion of the corresponding trellis graph with $5$ trees, where tree $13$ exhibits a fork structure.}

\section{Experiments}
    \label{sec:exp}

We use a Velodyne Puck VLP-16 lidar in our experiments. \textcolor{black}{The frequency of each lidar sweep is 5Hz. For our method, we run semantic lidar odometry (Alg.~\ref{alg:semantic_lidar_odometry}) $4\times$ per sweep, and semantic lidar mapping once per sweep.} The lidar is equipped onto a UAV platform that is manually flown, as well as a handheld sensor suite. The experiments were performed in Wharton State Forest in New Jersey. 

\begin{table}[]
\centering
\begin{tabular}{lll}
Layer  & Type & Filter \\\hline
1      & Downsampler block & $16$ \\
2-3    & $2x$ Non-bt-1D (no dilation) & $16$ \\
4      & Downsampler block & $32$ \\
5-6    & $2x$ Non-bt-1D (no dilation) & $32$ \\\hline
7      & Deconvolution & $32$ \\
8-11   & $4x$ Non-bt-1D (no dilation) & $32$ \\
12     & Deconvolution & Num. Classes
\end{tabular}
\caption{ERFNet inspired architecture used for semantic segmentation.}
\label{tbl:architecture}
\end{table}

\begin{figure}
    \begin{subfigure}{\columnwidth}
        \includegraphics[width=\columnwidth, trim={0cm 0cm 0cm 0cm},clip]{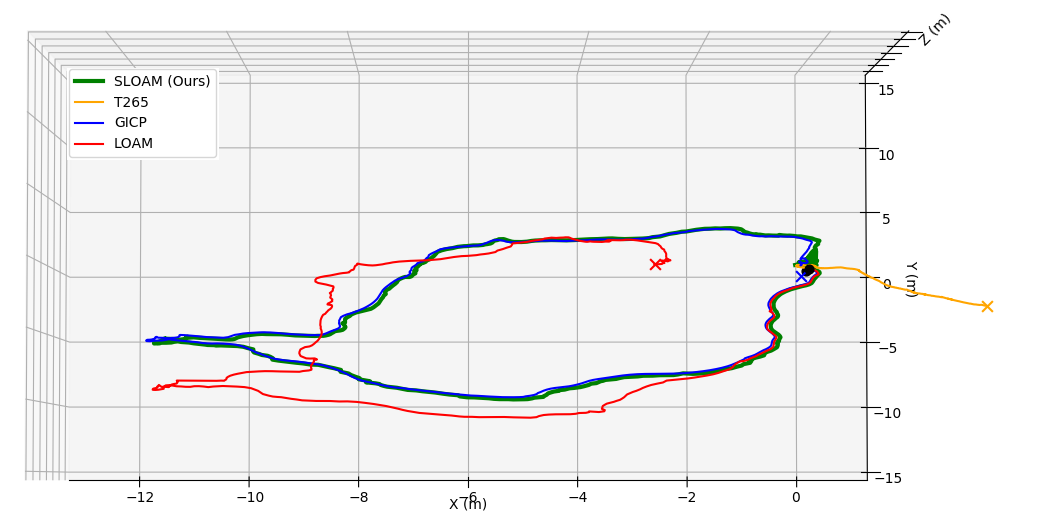}
    \end{subfigure}
    ~
    \begin{subfigure}{\columnwidth}
        \includegraphics[width=\columnwidth, trim={0cm 0cm 0cm 0cm},clip]{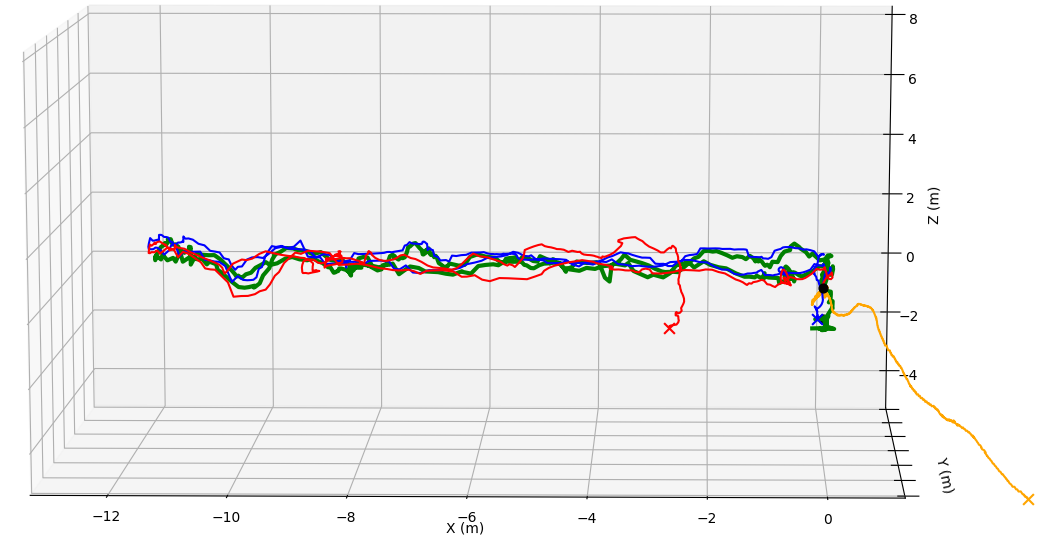}
    \end{subfigure}
    \caption{Trajectories of benchmark methods in hard UAV experiment loop.}
    \label{fig:trajectories}
\end{figure}


\begin{table}[]
\centering
\begin{tabular}{lcc}
Method & Distance from the goal (m) & Error \\\hline
Ours &  $0.37$ &  $0.58\%$\\
GICP &  $0.41$ &  $0.63\%$ \\
A-LOAM  &  $2.75$ &  $4.24\%$ \\
T265 (VIO) & $>100$ &  $>100\%$
\end{tabular}
\caption{Error: Distance relative to trajectory length in hard UAV experiment loop.}
\label{tbl:trajectory_error}
\end{table}

We benchmark our method (SLOAM) against A-LOAM, Generalized ICP (GICP), and the Intel Realsense T265. A-LOAM is an open source implementation of LOAM~\cite{zhang2014loam}. GICP~\cite{segal2009generalized} is an open source implementation of the Iterative Closest Point (ICP) algorithm available through PointCloud Library (PCL). \textcolor{black}{In order to increase the speed of ICP, we apply a voxel grid filter to reduce the number of points.} The Intel Realsense T265 is a commercial off-the-shelf tracking camera that relies on visual-inertial odometry (VIO).

We evaluate on 2 experiments, which we classify as medium and hard. The medium  scenario involves walking the handheld sensor suite through a dense forest environment for 1 minute in a straight line. The sensor readings entail rotation motions and instabilities caused by the operator dodging vines and thorns. The hard scenario is the UAV flying for 2 minutes that features significant rotation motions. It starts from hover and flies in a 65m trajectory until it loops back to the initial position and lands on a landing platform. Since the start and end marks are different in the $z$ axis, we offset the goal coordinate by 1 meter instead of using the origin.

\textcolor{black}{The segmentation network is trained on $544$ scans from which $16$ are from the handheld dataset and the remaining are from $5$ other regions of the same forest. No data from the UAV flight was used for training. The network architecture was chosen emphasizing inference speed. Our network architecture shown in Table~\ref{tbl:architecture} runs at 100Hz inference speeds with an Intel NUC5i7RYH on-board our UAV.  The final model achieves $0.81$ average IoU score on a 10-fold cross validation. }

We consider a variety of qualitative and quantitative metrics. Qualitatively we evaluate the trajectory and the accumulated point cloud to observe if any ghosting or duplication of trees occur. Quantitatively, we evaluate the error between start and end in the UAV experiment which performs a loop. Since SLOAM explicitly estimates the radius of each tree, we also quantitatively evaluate the DBH estimation compared to field measurements using a tape measure.

\begin{figure*}
    \begin{subfigure}{0.32\textwidth}
        \includegraphics[width=\columnwidth,trim={0cm 0cm 0cm 0cm},clip]{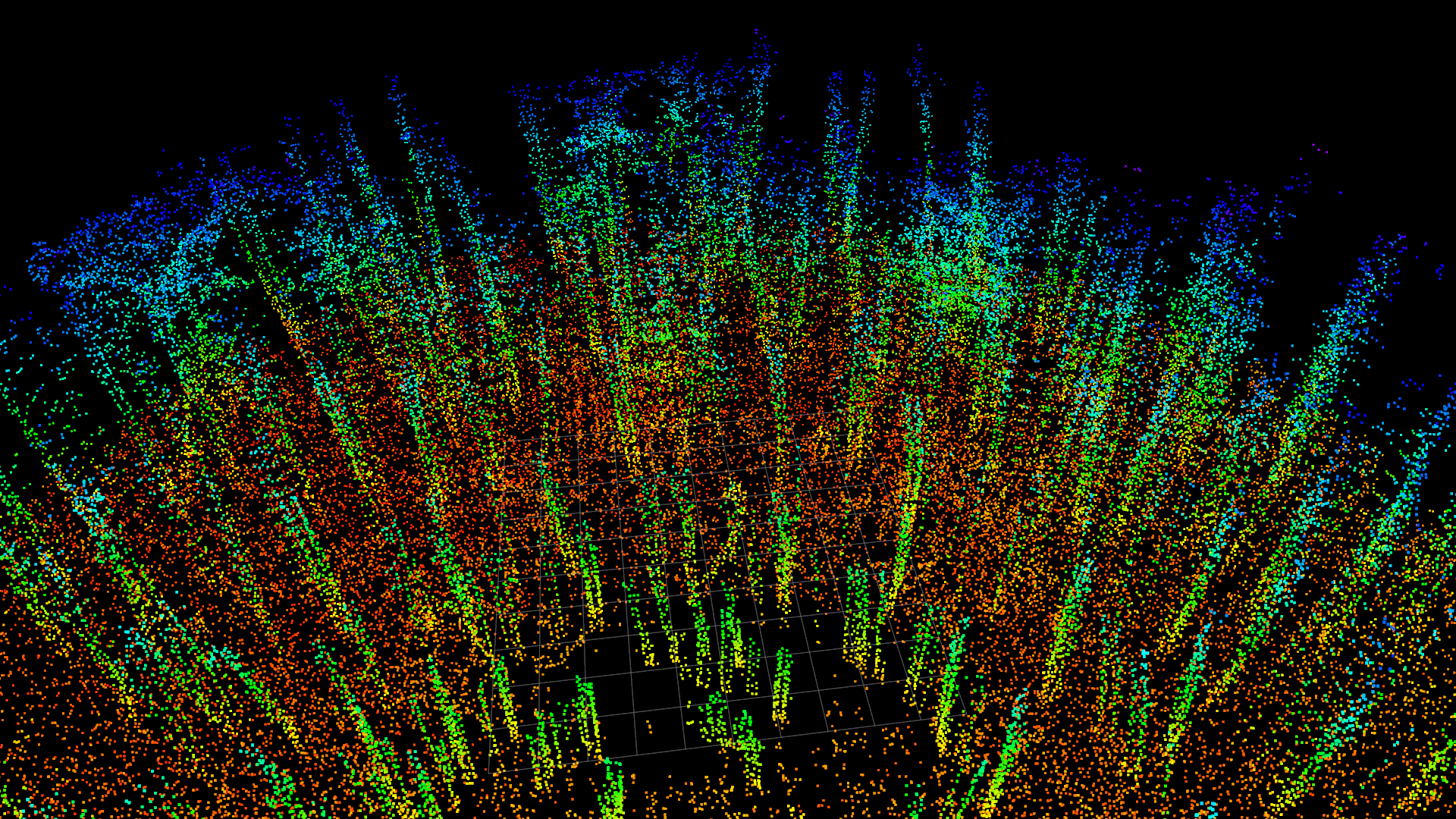}
        \vspace{0mm}
    \end{subfigure}
    ~
    \begin{subfigure}{0.32\textwidth}
        \includegraphics[width=\columnwidth,trim={0cm 0cm 0cm 0cm},clip]{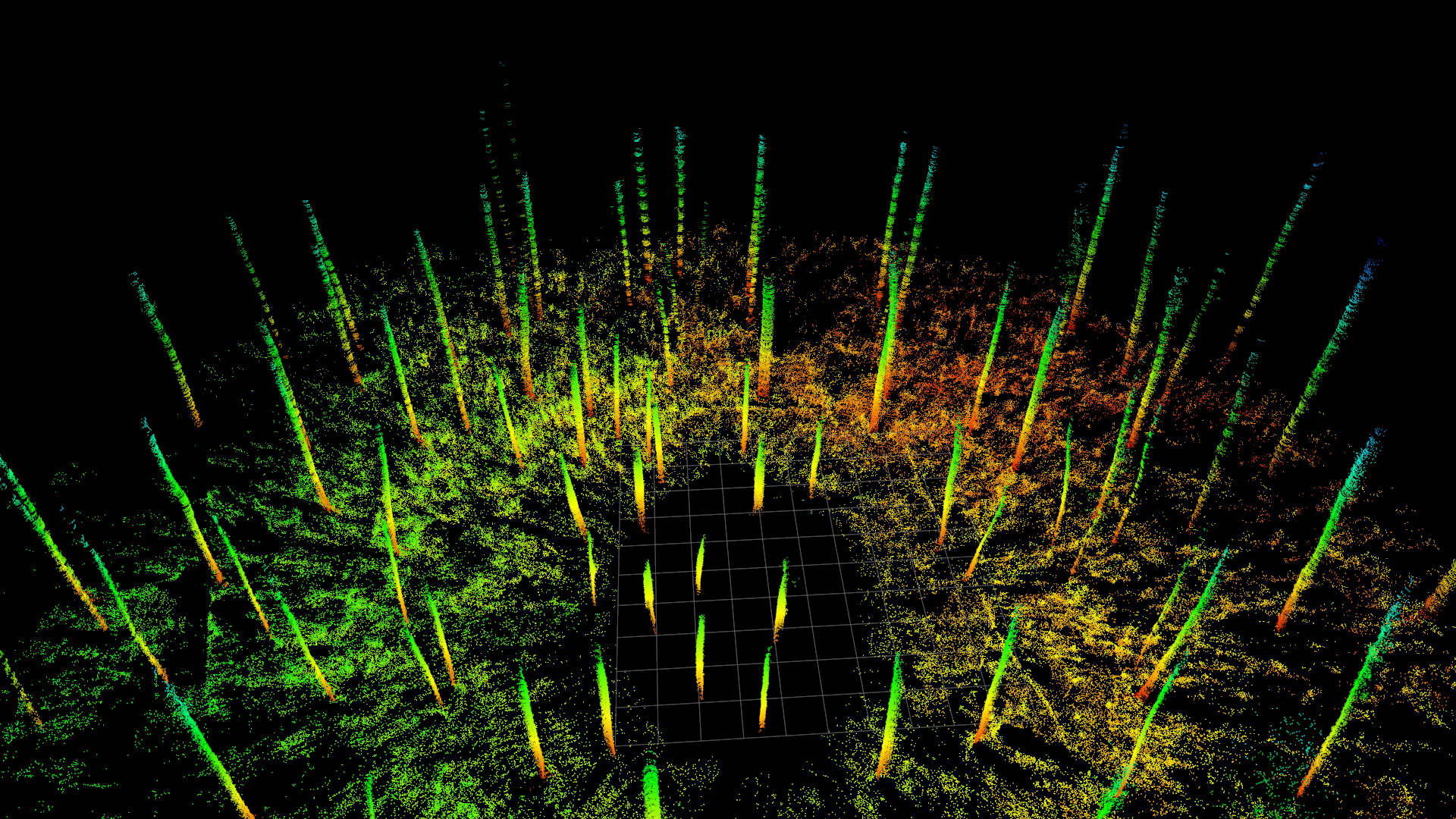}
        \vspace{0mm}
    \end{subfigure}
    ~
    \begin{subfigure}{0.32\textwidth}
        \includegraphics[width=\columnwidth, trim={0cm 0cm 0cm 0cm},clip]{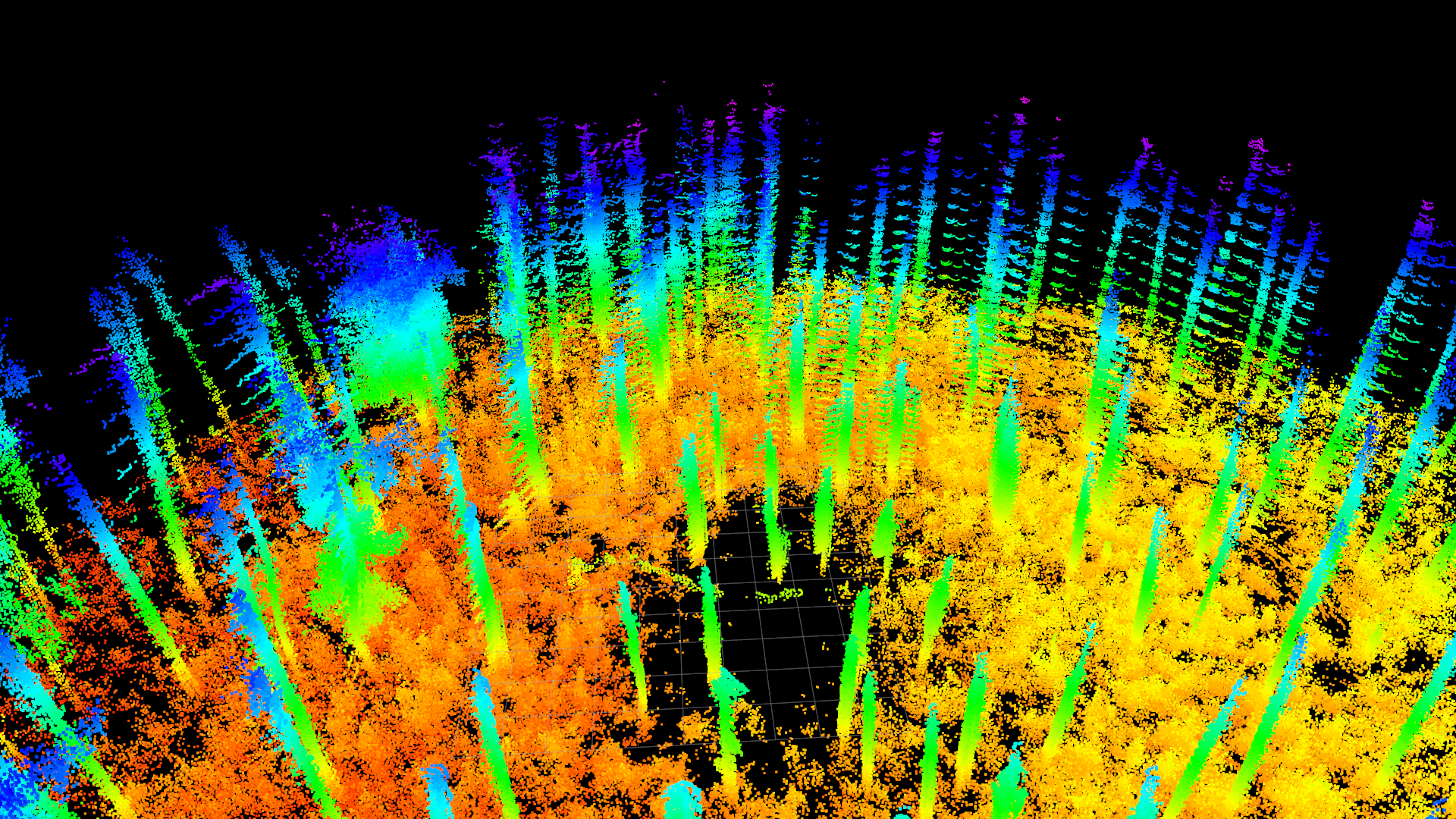}
        \vspace{0mm}
    \end{subfigure}
    ~
        \begin{subfigure}{0.32\textwidth}
        \includegraphics[width=\columnwidth,trim={0cm 0cm 0cm 0cm},clip]{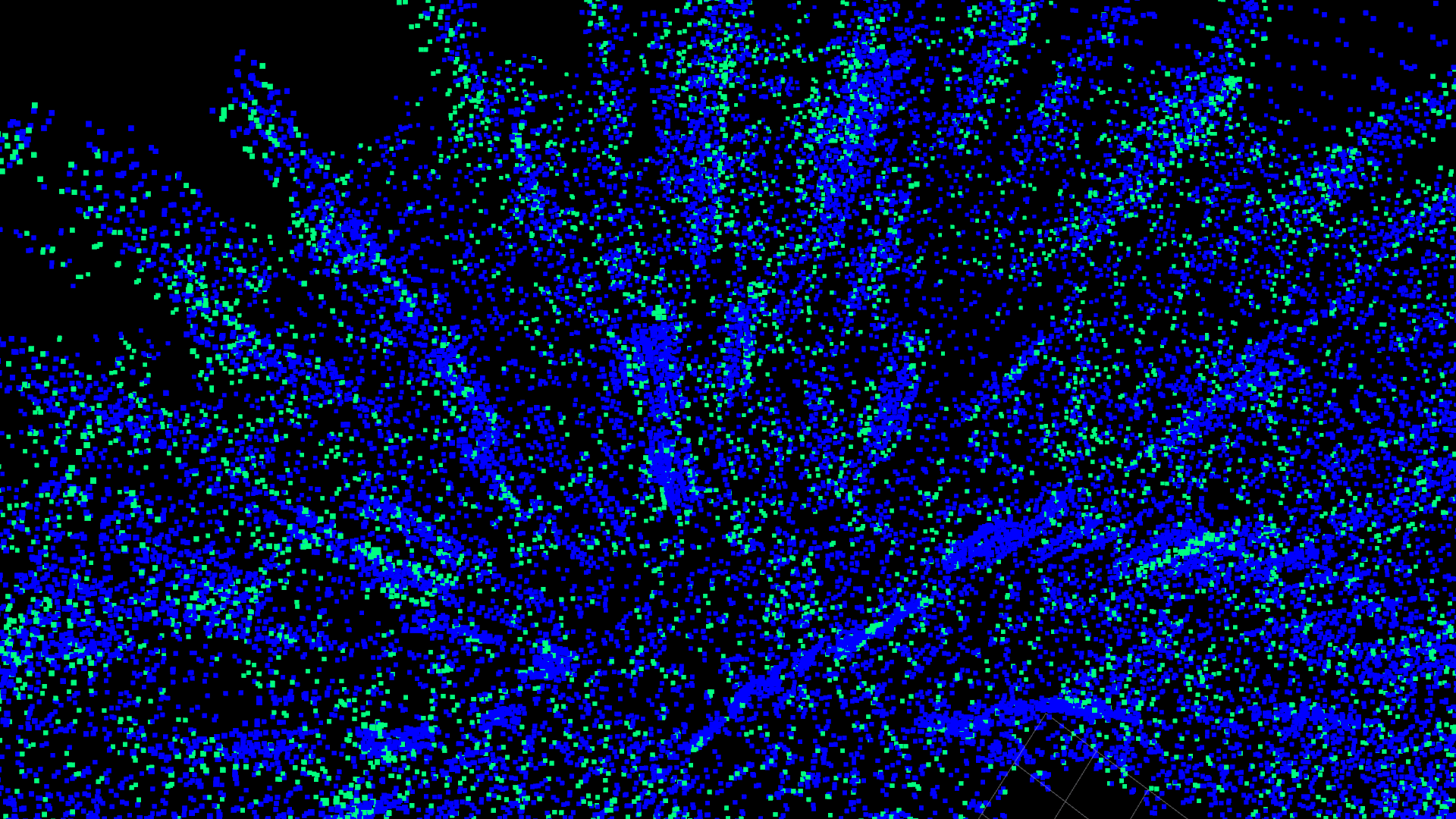}
        \caption{A-LOAM: Line and plane feature points are green (bottom).}
        \label{sfig:berry_loam}
    \end{subfigure}
    ~
    \begin{subfigure}{0.32\textwidth}
        \includegraphics[width=\columnwidth,trim={0cm 0cm 0cm 0cm},clip]{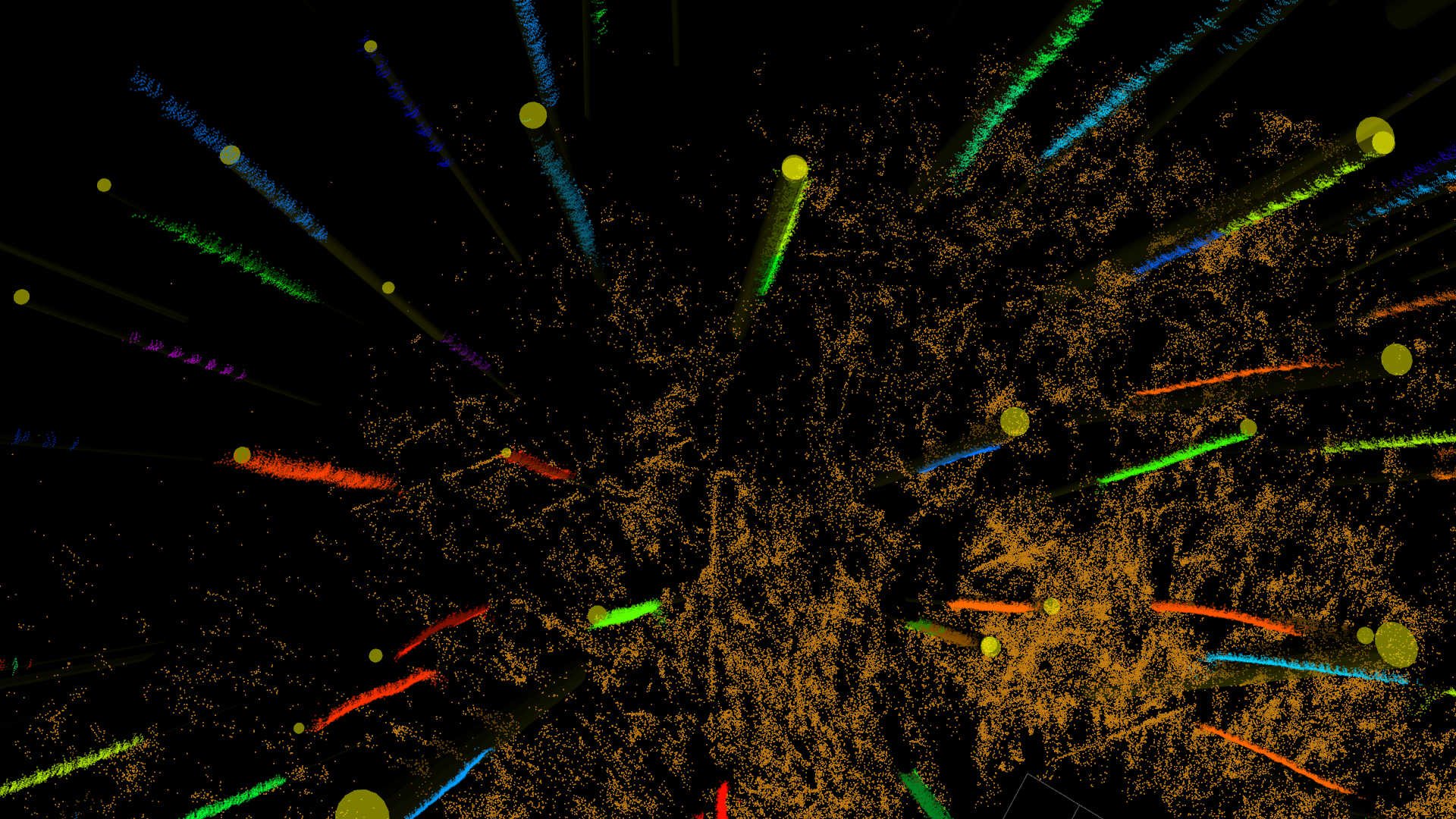}
        \caption{SLOAM (Our Method): Explicitly detects and models semantic landmarks.}
        \label{sfig:berry_sloam}
    \end{subfigure}
    ~
    \begin{subfigure}{0.32\textwidth}
    
        \includegraphics[width=\columnwidth, trim={0cm 0cm 0cm 0cm},clip]{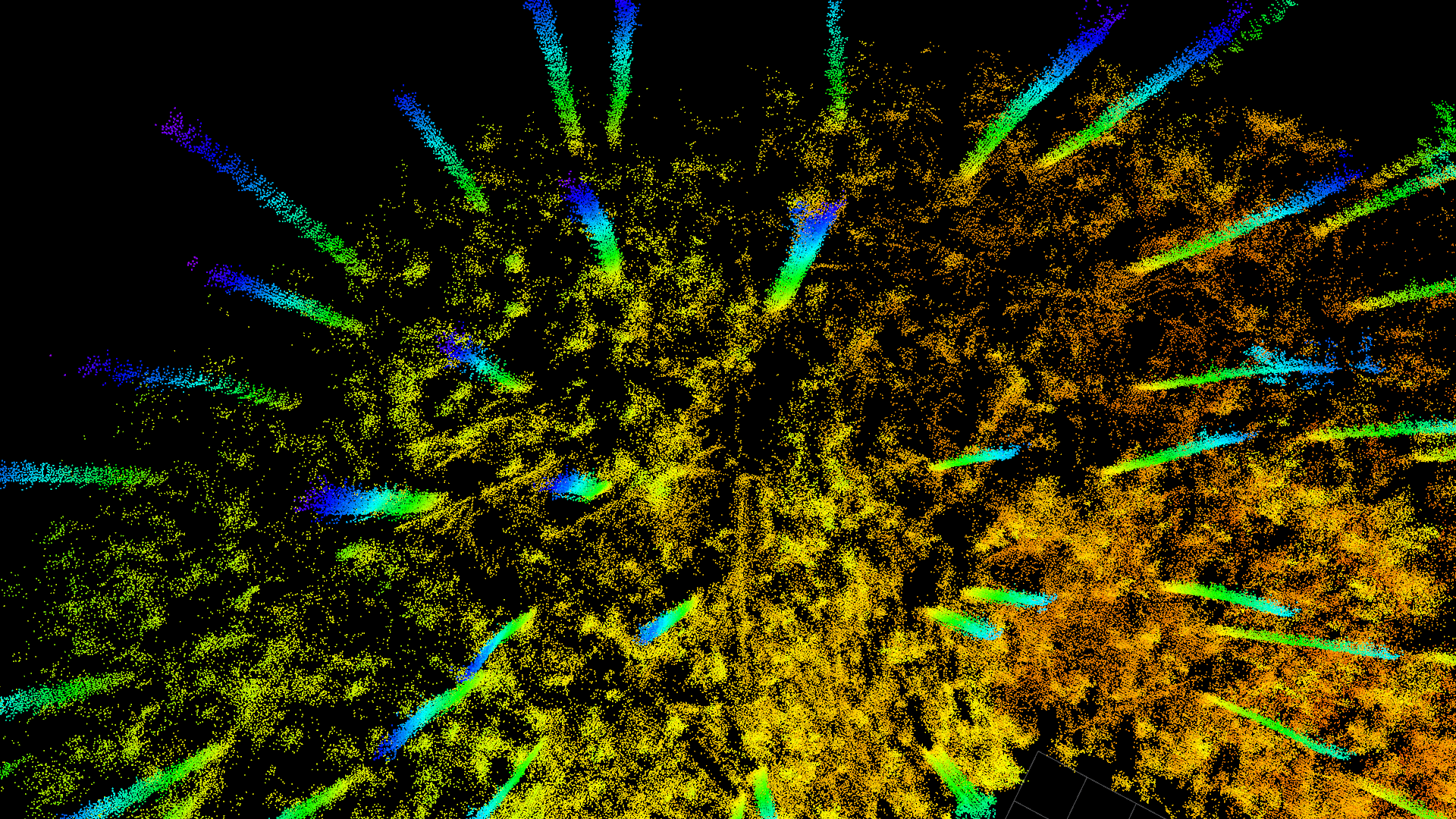}
        \caption{GICP: No distinction made between points.}
        \label{sfig:berry_icp}
    \end{subfigure}
    \caption{\textcolor{black}{SLOAM is the only method succeeds in handling aggressive motions (yaw). \textbf{Top Row} (UAV dataset): Colored by z-axis. A-LOAM and GICP maps are blurry, indicating failure in mapping. \textbf{Bottom Row} (Handheld dataset): Colors illustrate the different ways each method treats the points.}}
    \label{fig:biggrid}
    \vspace{-5mm}
\end{figure*}

\textcolor{black}{
Fig.~\ref{fig:trajectories} displays each method's trajectory on the UAV dataset, and Table~\ref{tbl:trajectory_error} quantifies the drift error between start and end. SLOAM achieves the lowest drift. A-LOAM clearly drifts, while the T265 outright fails. The VIO failure is expected, as it cannot handle extreme rotations. According to these trajectory metrics and plots, GICP seems to track closely with SLOAM, and achieves a similar magnitude of drift.}

\textcolor{black}{
However, Fig.~\ref{fig:biggrid} demonstrates that GICP also has difficulty with these datasets. The top row demonstrates the differences in the point cloud maps for each of the lidar based methods. Both A-LOAM and GICP produce blurry maps with frequent ghosting of trees. This blurring is unacceptable when we need to estimate the diameter of the trees with high accuracy. SLOAM, on the other hand, produces a crisp map that preserves fine details in the tree shapes. The fact that GICP performs comparably to SLOAM in the trajectory metrics, but much worse when viewing the point cloud maps, indicates that it has difficulty with rotation motions such as yaw, since the trajectory only measures $x$,$y$, and $z$ positions.}

\textcolor{black}{SLOAM outperforms A-LOAM and GICP because our semantic features are more reliable than texture-based lines and planes. Specifically, for both ground and tree features, data association is more robust since it inherently filters out noise, and the resulting cost function is more informative due to the use of landmark shapes. While these texture features are reliable in man-made environments, they are problematic in natural environments which lack clear planar and edge surfaces. The bottom row of Fig.~\ref{fig:biggrid} illustrates the different ways each method treats the points. SLOAM detects each semantic landmark, indicated by the different colorings of each tree instance. On the other hand, the A-LOAM features appear random, indicating that they are not distinctive and are thus prone to frequent data misassociation. Finally, GICP does not make a distinction between the points. This approach works well when the motion is slight, but it is also susceptible to data misassociation during extreme rotations.}

\textcolor{black}{Compared to SLOAM which uses a point to cylinder cost function, both A-LOAM and GICP only utilize a point to plane or point to line cost functions to compute the pose transformation. These approaches force a false planar model onto the cylinders, and will introduce slight errors that manifest as wider, blurry trees. While these small errors will not lead to an outright failure in the sensor state estimates, they are still unacceptable due to the high precision and accuracy necessary to measure tree diameters.
}

\begin{table}[]
\centering
\begin{tabular}{c|cccc}
    Detected Trees & Mean & Median  & Max & Min \\\hline
    $29$ &$0.67$ & $0.6$  & $1.4$ & $0.1$
\end{tabular}
\caption{DBH Metrics in hard UAV experiment}
\label{tbl:DBH}
\end{table}

We next evaluate how well SLOAM can estimate the DBH of the tree landmarks. We obtain these estimate for free, as we can use the semantic models and features to extract out the diameter for each landmark. For the hard UAV dataset described in the previous section, we manually measured $35$ trees that were in the path of the robot and use the models generated by our method to estimate DBH. We summarize the DBH estimation results compared to human measurements in Table~\ref{tbl:DBH}. SLOAM detected $29$ trees with an average error is $0.67$ in, which falls within the desired accuracy as typically in industry the measurements are taken to the nearest inch. 


We found that using the diameter parameter of our cylinder models to estimate the DBH had a few large outliers. Instead, we found that it was more effective to take the median of all the radii estimates across all beams in all scans, which yielded the results presented in Table~\ref{tbl:DBH}. This process still requires accurate registration from SLOAM to group these beams together. We believe that since only half of the cylinder can be viewed from the lidar at a single scan, the presence of noise or insufficient features can cause instability in the geometric least squares optimization process. It does not seem to affect the pose estimation optimization process, as the presence of many tree landmarks offers robustness to noise. \textcolor{black}{Beyond the median operation, no additional post-processing steps are required to obtain the DBH results.}

The ability to quantify mapping performance through landmark ground truth measurements highlights another strength of our approach. It is difficult to quantify the performance of traditional SLAM algorithms, as ground truth measurements of trajectories are hard to obtain in natural environments. \textcolor{black}{For example, the dense forest canopy prevents the use of GPS as ground truth, as the errors can range up to tens of meters}. On the other hand, ground truth measurements of the landmark shapes are easily obtained, and provide an alternative quantitative metric to benchmark various algorithms.




\section{Conclusions}
    \label{sec:conc}
    We pose the DBH estimation problem as a special case of the semantic lidar odometry and mapping problem, providing a generic formulation that can be extended to other scenarios. We develop a VR labeling tool to facilitate annotation of 3D lidar scans by fitting geometric primitives to the raw data. This enables us to train a segmentation model that is used with a graph based method for instance detection and extraction of relevant attributes of each tree such as radius and focus point. We demonstrate the difficulty of the DBH estimation problem by benchmarking against $3$ other lidar state estimation and VIO methods and observing large drifts. Finally, we show that the semantic shape models are critical in achieving accuracy and scalability in challenging natural environments. 

\bibliographystyle{./bibliography/IEEEtran}
\bibliography{./bibliography/references}

\end{document}